\documentclass[conference]{IEEEtran}

\usepackage{amsmath,amssymb}
\usepackage{graphicx}
\usepackage{booktabs}
\usepackage{multirow}
\usepackage{url}
\usepackage{hyperref}
\usepackage{xcolor}
\usepackage{enumitem}
\usepackage{subcaption}
\usepackage{balance}

\hypersetup{colorlinks=true,linkcolor=black,citecolor=black,urlcolor=blue}
\begin{document}

\title{AT-Attn: Temporal-Aware Cross-Attention for\\Longitudinal Multimodal Alzheimer's Disease Diagnosis}

\author{
\IEEEauthorblockN{Xinyue Du\IEEEauthorrefmark{1},
Yibo Liu\IEEEauthorrefmark{2},
Zhenglei Zhou\IEEEauthorrefmark{3},
Xuancheng Yao\IEEEauthorrefmark{2},
Weimin Zhong\IEEEauthorrefmark{1},
Qiuhui Chen\IEEEauthorrefmark{1}\IEEEauthorrefmark{4}}
\IEEEauthorblockA{\IEEEauthorrefmark{1}School of Information Science and Engineering,
East China University of Science and Technology, China\\
Email: xiaoduxiaodu09@gmail.com}
\IEEEauthorblockA{\IEEEauthorrefmark{2}School of Computer Science,
Shanghai Jiao Tong University, China}
\IEEEauthorblockA{\IEEEauthorrefmark{3}Tencent, China}
\IEEEauthorblockA{\IEEEauthorrefmark{4}Corresponding author: Qiuhui Chen (chenqh@ecust.edu.cn)}
}

\maketitle

\begin{abstract}
In longitudinal Alzheimer's disease (AD) diagnosis support, clinical and imaging information is often collected at irregular visits. Integrating these multimodal observations may improve diagnostic assessment, but naive fusion can degrade performance when MRI is noisy or intermittently unavailable. We propose AT-Attn, a temporal-aware multimodal framework that combines Change-and-Time encoding, time-biased asymmetric cross-attention, and gated fusion to integrate MRI with longitudinal clinical information. We evaluate AT-Attn on an MRI-retained ADNI cohort of 1,520 patients using structural MRI, six cognitive-scale trajectories, and seven static clinical variables under patient-level five-fold cross-validation. The main asymmetric AT-Attn model achieves accuracy 0.719$\pm$0.024, macro F1 0.721$\pm$0.023, ROC-AUC 0.873$\pm$0.013, and PR-AUC 0.783$\pm$0.018, outperforming unimodal and naive multimodal fusion baselines while remaining competitive with strong tabular baselines. These results suggest that a temporal-aware and constrained fusion strategy can help structural MRI contribute clinically relevant complementary information for patient-level AD diagnosis support.
\end{abstract}


\begin{figure*}[!t]
\centering
\includegraphics[width=0.95\textwidth]{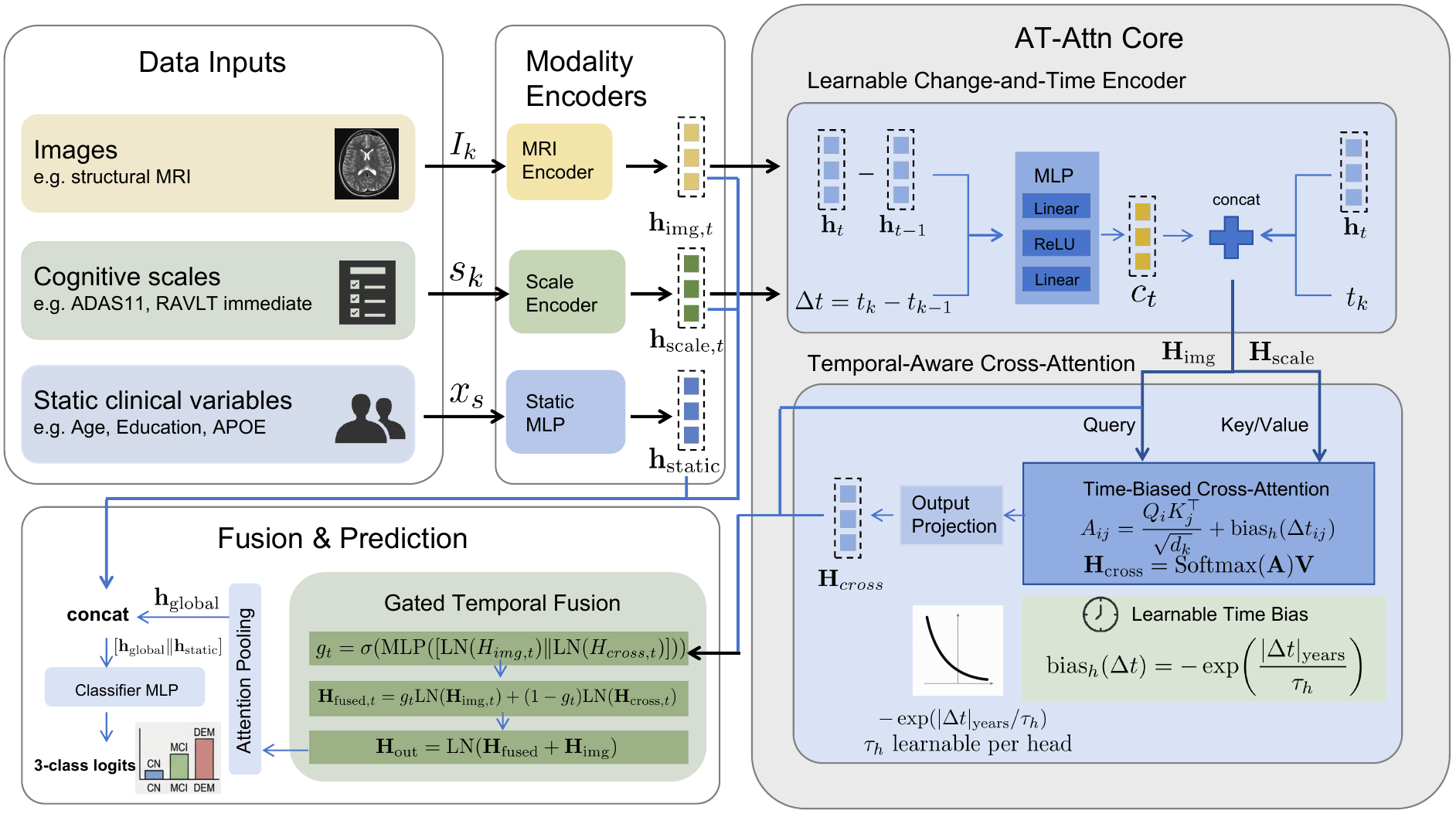}
\caption{Overview of the AT-Attn framework. MRI and cognitive-scale features are first encoded and enriched by modality-specific Change-and-Time encoders. MRI then queries cognitive-scale representations through time-biased asymmetric cross-attention, and the resulting representations are integrated by gated fusion. The fused sequence and complementary direct representations are temporally pooled and combined with static clinical features for final classification.}
\label{fig:arch}
\end{figure*}

\section{Introduction}
\label{sec:intro}

Alzheimer's disease (AD) is the leading cause of dementia and is characterized by a progressive transition from cognitively normal (CN) status to mild cognitive impairment (MCI) and eventually dementia~\cite{alz2025}. Accurate identification of MCI and current disease stage is clinically important because it can support timely intervention, longitudinal monitoring, and more informed patient management~\cite{petersen1999,albert2011mci}. In both routine clinical workflows and longitudinal cohort studies such as the Alzheimer's Disease Neuroimaging Initiative (ADNI), patient assessment is inherently multimodal~\cite{petersen2010}. Structural MRI captures brain atrophy and neurodegenerative patterns~\cite{vemuri2010}, cognitive and functional assessments reflect clinical impairment, and demographic or genetic factors such as APOE-$\varepsilon$4 provide complementary risk information~\cite{duara1996}. Developing computational methods that can effectively combine these heterogeneous signals therefore remains an important biomedical informatics problem.

However, the multimodal setting relevant to practice is not simply ``more modalities are better.'' In AD cohorts, cognitive assessments are clinically direct, whereas structural MRI is high-dimensional, noisier, and often unavailable at some visits. MRI may provide complementary information, but naive fusion can also degrade performance by allowing the weaker modality to distort stronger clinical signals. For longitudinal AD diagnosis, this problem is further complicated by irregular follow-up intervals, meaning that cross-modal interaction should account for temporal proximity~\cite{baytas2017tlstm,che2018,rubanova2019}. These observations motivate not only temporal-aware fusion, but a constrained fusion mechanism that can inject MRI information in a stable and complementary way while preserving direct clinical evidence.

To address this problem, we propose \textbf{AT-Attn} (\textbf{T}emporal-\textbf{A}ware Cross-\textbf{Att}e\textbf{n}tion), a longitudinal multimodal framework for patient-level AD diagnosis support that combines modality-specific encoding, Change-and-Time encoding, time-biased asymmetric cross-attention, and gated fusion. AT-Attn lets MRI representations query cognitive-scale representations under an explicit temporal penalty, enabling cross-modal interaction while preserving unimodal evidence.

The main contributions of this work are as follows:
\begin{itemize}[nosep,leftmargin=*]
  \item We study longitudinal multimodal AD diagnosis from the perspective of \textbf{stable MRI integration}, aiming to extract complementary MRI value without degrading stronger clinical signals under irregular follow-up.
  \item We introduce a \textbf{temporal-aware multimodal architecture} that models longitudinal MRI and cognitive-scale dynamics, performs asymmetric fusion with additive temporal bias and gating, and incorporates static clinical context.
  \item We evaluate the framework on an MRI-retained ADNI cohort with patient-level five-fold cross-validation, reproduced baselines, systematic ablations, and subgroup analyses, showing gains over unimodal and naive multimodal fusion baselines, competitive tabular performance, and context-dependent MRI contribution.
\end{itemize}

\section{Related Work}
\label{sec:related}

\subsection{Clinical and Biomarker Background}

AD is now widely viewed as a biological and clinical continuum, with mild cognitive impairment (MCI) representing an important intermediate stage between normal cognition and dementia~\cite{petersen1999,albert2011mci,sperling2011preclinical}. Conceptual biomarker models further suggest temporally ordered but overlapping changes in amyloid, tau, neurodegeneration, and cognition across the disease course~\cite{jack2010cascade,jack2013tracking,dubois2016preclinical,jack2016atn}. Within this continuum, structural MRI, PET, cerebrospinal fluid biomarkers, and cognitive assessments each capture complementary aspects of disease state and progression~\cite{vemuri2010,ewers2011markers,zandifar2020}, which motivates multimodal computational approaches for diagnosis and prognosis.

\subsection{Multimodal AD Diagnosis}

Multimodal AD diagnosis research has progressed from classical feature-based fusion to deep representation learning. Early studies showed that combining MRI, PET, and CSF can outperform single-modality models for AD and MCI classification~\cite{zhang2011multimodal,zhang2011multitask}, and later work expanded this direction with multimodal feature selection and deep fusion strategies~\cite{suk2013deepfeature,suk2014hierarchical,liu2015multiclass,zhu2016canonical,zu2016label,kim2018msh}. More recent studies have explored MRI-based deep models, explainable multimodal systems, and richer fusion with PET, diffusion MRI, cognitive variables, and metadata~\cite{basaia2019,qiu2020,elsappagh2021xai,li2025diamond,huang2025multistage,li2025agnflow,yin2025unicross}. Reviews consistently conclude that multimodal information is beneficial, although comparisons remain confounded by differences in cohorts, label definitions, and evaluation protocols~\cite{rathore2017review,pellegrini2018review}.

\subsection{Longitudinal AD Modeling}

Beyond cross-sectional diagnosis, several studies have modeled disease progression or MCI-to-AD conversion from longitudinal biomarkers. Representative approaches include multitask longitudinal regression over multimodal biomarkers~\cite{tabarestani2020distributed} and deep recurrent forecasting models for future diagnosis and cognition~\cite{nguyen2020deeprnn}. Clinical studies have also shown that MRI and cognitive scores provide complementary longitudinal prognostic information years before dementia onset~\cite{zandifar2020}. However, most prior AD-specific longitudinal models either concatenate modalities directly or summarize them into a shared latent state, without explicitly controlling how weaker and stronger modalities should interact when cognitive information is limited.

\subsection{Irregular Temporal Modeling and Multimodal Robustness}

Irregular sampling has been widely studied in clinical time-series modeling through time-aware recurrent networks, imputation-based models, continuous-time dynamics, and attention-based encoders~\cite{baytas2017tlstm,neil2016phased,che2018,alolaimat2024,cao2018brits,rubanova2019,debrouwer2019gruode,kidger2020ncde,shukla2021mtan}. However, these methods mainly address within-modality temporal irregularity rather than how cross-modal interaction should be modulated when visit intervals are uneven and modalities contribute unequal predictive signal.

In parallel, multimodal learning research has emphasized representation mismatch, missing modalities, and optimization imbalance~\cite{baltrusaitis2019survey,bayoudh2022survey,wang2020makes,peng2022balanced}. These concerns are especially relevant in AD, where cognitive scores can dominate imaging features and longitudinal cohorts often contain incomplete MRI follow-up~\cite{li2025agnflow,yin2025unicross}. Our work is positioned at the intersection of these two threads, targeting patient-level diagnosis-support classification under irregular longitudinal follow-up while accounting for unequal cross-modal predictive strength.

\section{Method}
\label{sec:method}

\subsection{Problem Formulation}

Each patient is represented as a sequence of study visits $k = 1,\ldots,T$ with shared calendar timestamps $t_k$ (months from baseline). At each visit we have: MRI volume $I_k \in \mathbb{R}^{96\times112\times96}$, cognitive scale vector $s_k \in \mathbb{R}^{6}$, and static patient features $\mathbf{x}_{\text{static}} \in \mathbb{R}^{7}$. The patient-level label $y \in \{\text{CN}, \text{MCI}, \text{DEM}\}$ is defined by the diagnosis at the last retained visit on the shared visit grid, so the task is patient-level disease-stage classification using the retained longitudinal record. Visit intervals are irregular, with spacing that varies substantially across patients.

Some visits on the shared longitudinal grid do not have matched MRI scans. We therefore associate each visit with an MRI availability indicator
\begin{equation}
m^{\text{MRI}}_k \in \{0,1\},
\end{equation}
where $m^{\text{MRI}}_k = 1$ denotes that a valid MRI is available at visit $k$. This mask is used by downstream modules to identify valid MRI time steps and ignore unmatched image visits.

\subsection{Overall Architecture}

Fig.~\ref{fig:arch} shows the revised AT-Attn framework. The main pipeline comprises five stages: (1)~modality-specific feature encoding, (2)~Change-and-Time encoding, (3)~temporal-aware asymmetric cross-attention, (4)~gated fusion, and (5)~temporal pooling and classification. A 3D ResNet-18 backbone, initialized with Med3D~\cite{chen2019med3d} pretrained weights, encodes each MRI frame and projects it to a unified dimension $D = 256$ via a linear layer followed by LayerNorm and ReLU. Cognitive-scale features undergo an analogous projection to $\mathbb{R}^{D}$, while static variables are encoded by a separate MLP.

Both longitudinal branches are then processed by independent Change-and-Time encoders (Sec.~\ref{sec:cre}) before asymmetric cross-attention (Sec.~\ref{sec:atattn}) refines the MRI branch using cognitive-scale-derived context. The cross-attended MRI sequence is fused with the original MRI sequence through a gate (Sec.~\ref{sec:gate}). In parallel, direct MRI and cognitive-scale representations are also pooled through shortcut paths to preserve unimodal evidence. The classifier finally receives the concatenation of pooled direct MRI features, pooled fused MRI features, pooled direct cognitive-scale features, and static features.

\subsection{Learnable Change-and-Time Encoder}
\label{sec:cre}

AD pathology is progressive, so longitudinal change between visits can be informative for distinguishing stable impairment from ongoing decline. Let $\mathbf{h}_{\text{img},t}, \mathbf{h}_{\text{scale},t} \in \mathbb{R}^{D}$ denote the modality-specific encoded features at time step $t$. Direct temporal differencing $\Delta F / \Delta t$ can be numerically unstable when $\Delta t$ varies from months to years. Rather than explicitly computing temporal derivatives, we encode three complementary signals per time step:
\begin{align}
\Delta \mathbf{h}_t &= \mathbf{h}_t - \mathbf{h}_{t-1},\quad \Delta t_{\text{norm}} = \frac{\max(\Delta t, 1.0)}{12.0} \label{eq:delta_rev}\\
\mathbf{c}_t &= \text{MLP}_{\text{change}}([\Delta \mathbf{h}_t \| \Delta t_{\text{norm}}]) \label{eq:change_rev}\\
\mathbf{e}_t &= \text{MLP}_{\text{time}}(t/12.0) \label{eq:tembed_rev}\\
\mathbf{H}_t &= \text{LN}(\text{Linear}([\mathbf{h}_t \| \mathbf{c}_t \| \mathbf{e}_t])) \label{eq:cre_out_rev}
\end{align}
where $\mathbf{H}_t \in \mathbb{R}^{D}$ is the enriched representation after Change-and-Time encoding. The three components encode the instantaneous feature state, inter-visit feature differences together with elapsed time, and the absolute temporal position. This design allows the network to learn time-aware change representations without requiring an explicit division by $\Delta t$. MRI and cognitive-scale branches follow the same formulation but use separate learnable parameters to capture modality-specific temporal dynamics.

When MRI is missing at visit $t$, the corresponding mask value $m^{\text{MRI}}_t = 0$ suppresses transitions involving that time step so that neither zero-filled images nor missing-to-valid transitions contribute artificial change signals.

\subsection{Temporal-Aware Cross-Attention}
\label{sec:atattn}

After Change-and-Time encoding, we obtain $\mathbf{H}_{\text{img}}, \mathbf{H}_{\text{scale}} \in \mathbb{R}^{B \times T \times D}$. We adopt an asymmetric formulation in which MRI features act as queries and cognitive-scale features as keys/values:
\begin{equation}
\mathbf{Q} = \mathbf{H}_{\text{img}}\mathbf{W}_Q,\;
\mathbf{K} = \mathbf{H}_{\text{scale}}\mathbf{W}_K,\;
\mathbf{V} = \mathbf{H}_{\text{scale}}\mathbf{W}_V.
\end{equation}

This asymmetric direction is motivated by the imbalance in modality strength in our setting, where cognitive scales are often more directly diagnostic than MRI. We therefore use cognitive-scale-derived context to refine the MRI branch, aiming to limit noisy MRI-derived feedback into the stronger clinical branch. This conservative design is intended to preserve reliable unimodal evidence while still enabling cross-modal interaction~\cite{wang2020makes,peng2022balanced,yin2025unicross}.

\textbf{Time bias.} We augment scaled dot-product attention with a learnable additive temporal penalty. For each pair of time steps $i, j \in \{1,\ldots,T\}$,
\begin{align}
A_{ij} &= \frac{\mathbf{Q}_i \mathbf{K}_j^\top}{\sqrt{d_k}} + \text{bias}(|t_i - t_j|) \label{eq:attn_rev}\\
\text{bias}(\Delta t) &= -\exp\!\left(\frac{|\Delta t|_{\text{years}}}{\tau_h}\right),\quad \tau_h > 0 \text{ per head}. \label{eq:bias_rev}
\end{align}
Because this bias is added to the pre-softmax logits, $-\exp(\Delta t/\tau_h)$ explicitly down-weights temporally distant keys, whereas $\exp(-\Delta t/\tau_h)$ mainly gives nearby visits a small positive bonus and approaches 0 for distant ones. On our harmonized visit grid ($\Delta t\in[0,6]$ years), the resulting penalty stays within a bounded range, so it modulates attention ranking without behaving like a hard mask. Empirically, the adopted form performed better under the same five-fold protocol.

To ensure positivity, each head-specific $\tau_h$ is parameterized as
\begin{equation}
\tau_h = \operatorname{softplus}(\hat{\tau}_h) + \epsilon,\qquad \epsilon = 10^{-3},
\end{equation}
where $\hat{\tau}_h\in\mathbb{R}$ is learnable. Padding and unavailable MRI query positions are masked before softmax. The output is $\mathbf{H}_{\text{cross}} = \text{Softmax}(\mathbf{A}) \cdot \mathbf{V}$, followed by linear projection, dropout, and LayerNorm with a residual connection.

\subsection{Gated Fusion, Temporal Pooling, and Classification}
\label{sec:gate}

Rather than replacing the MRI representation with the cross-attention output, we use a per-time-step gate to balance raw and cross-attention-enhanced features:
\begin{align}
g_t &= \sigma\Bigl(\text{MLP}([\text{LN}(\mathbf{H}_{\text{img},t}) \| \notag\\
&\qquad \text{LN}(\mathbf{H}_{\text{cross},t})])\Bigr) \in [0,1] \label{eq:gate_rev}\\
\mathbf{H}_{\text{fused},t}
&= g_t \cdot \text{LN}(\mathbf{H}_{\text{img},t}) \notag\\
&\quad + (1-g_t) \cdot \text{LN}(\mathbf{H}_{\text{cross},t}) \label{eq:fuse_rev}\\
\mathbf{H}_{\text{out}} &= \text{LN}(\mathbf{H}_{\text{fused},t} + \mathbf{H}_{\text{img},t}). \label{eq:out_rev}
\end{align}
The gate preserves flexibility in how much direct MRI evidence should be retained at each visit.

Temporal aggregation uses attention pooling:
\begin{equation}
\alpha_t = \frac{\exp(\mathbf{w}^\top \tanh(\mathbf{W}_a \mathbf{H}_{t}))}
{\sum_{t'} \exp(\mathbf{w}^\top \tanh(\mathbf{W}_a \mathbf{H}_{t'}))}. \label{eq:pool_rev}
\end{equation}
For MRI-derived sequences, pooling weights at unavailable MRI visits are masked out and the remaining valid weights are renormalized over observed visits only.
We apply the same pooling operator to the fused MRI sequence as well as to direct MRI and cognitive-scale sequences exposed through shortcut paths, producing
\begin{equation}
\mathbf{h}_{\text{fused}},\quad \mathbf{h}_{\text{img-direct}},\quad \mathbf{h}_{\text{scale-direct}} \in \mathbb{R}^{D}.
\end{equation}
These are concatenated with the encoded static-feature vector $\mathbf{h}_{\text{static}} \in \mathbb{R}^{32}$ to form
\begin{equation}
\mathbf{z} = [\mathbf{h}_{\text{img-direct}} \| \mathbf{h}_{\text{fused}} \| \mathbf{h}_{\text{scale-direct}} \| \mathbf{h}_{\text{static}}] \in \mathbb{R}^{3D+32}.
\end{equation}
With $D = 256$, the classifier input dimension is 800. This shortcut design exposes strong direct clinical evidence to the classifier while still allowing the fused branch to model complementary MRI--scale interaction.

\section{Experiments}
\label{sec:exp}

\subsection{Dataset and Implementation}

\begin{figure}[!t]
\centering
\includegraphics[width=0.98\columnwidth]{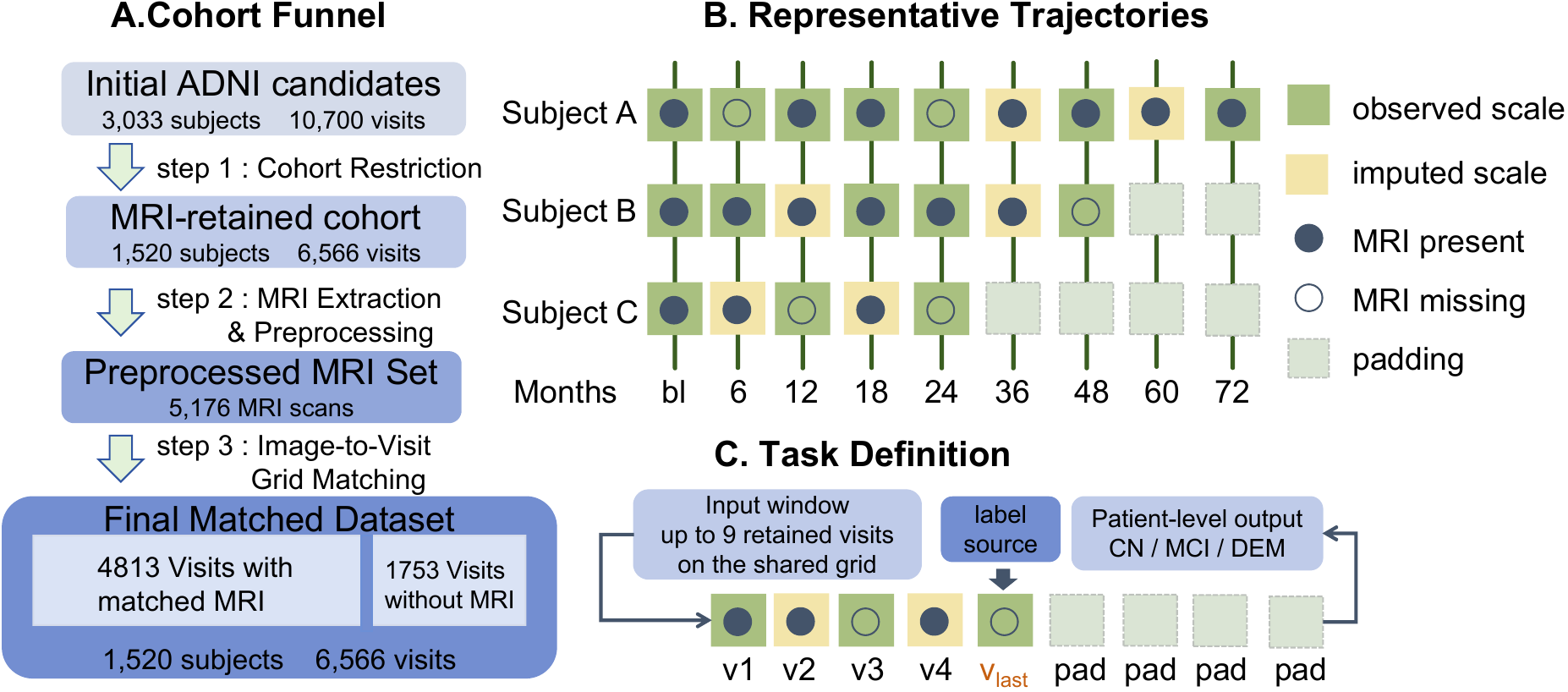}
\caption{Cohort overview and task setup. (A) Cohort construction. (B) Representative trajectories. (C) Patient-level staging setup.}
\label{fig:data_panel}
\end{figure}

\textbf{Dataset.} We use ADNI subjects with longitudinal structural MRI and corresponding clinical data~\cite{petersen2010}. For MRI-based experiments, we retain T1-weighted 3D structural MRI scans acquired at 3.0\,T with slice thickness $\leq 1.5$\,mm, yielding an MRI-retained cohort of 1,520 patients after VISCODE normalization and subject-visit matching. The longitudinal cognitive input consists of six scales: ADAS11, ADAS13, RAVLT immediate, RAVLT learning, RAVLT forgetting, and RAVLT percent forgetting. Static features include age at baseline, years of education, sex, ethnicity, race, marital status, and APOE-$\varepsilon$4 allele count. Table~\ref{tab:cohort} summarizes cohort statistics, and Fig.~\ref{fig:data_panel} provides a visual overview of cohort construction, representative longitudinal trajectories, and the task definition.

\begin{table}[!t]
\centering
\caption{Summary of cohort construction and MRI availability statistics. In the final row, MRI availability is reported as matched/missing visits on the shared visit grid.}
\label{tab:cohort}
\setlength{\tabcolsep}{4pt}
\small
\renewcommand{\arraystretch}{1.12}
\begin{tabular}{@{}p{3.15cm}ccp{1.75cm}@{}}
\toprule
Cohort step & Subjects & Visits & MRI availability \\
\midrule
Initial ADNI candidates & 3,033 & 10,700 & -- \\
\midrule
MRI-retained cohort for MRI-based experiments & 1,520 & 6,566 & -- \\
\midrule
Preprocessed MRI image set & 1,520 & -- & 5,176 \\
\midrule
Matched MRI on shared visit grid & 1,520 & 6,566 & 4,813 / 1,753 \\
\bottomrule
\end{tabular}
\end{table}

\begin{table*}[!t]
\centering
\caption{PERFORMANCE COMPARISON ON THE ADNI COHORT UNDER FIVE-FOLD CROSS-VALIDATION. THE BEST RESULTS ARE IN BOLD.}
\label{tab:main}
\setlength{\tabcolsep}{4pt}
\small
\begin{tabular}{@{}lccccc@{}}
\toprule
Method & Acc & Macro-F1 & ROC-AUC & PR-AUC & MCI Sens@90Spe / Achieved Spe \\
\midrule
LGBM & 0.7130$\pm$0.0212 & 0.7173$\pm$0.0180 & 0.8693$\pm$0.0132 & 0.7679$\pm$0.0268 & 0.3578$\pm$0.0721 / 0.8846$\pm$0.0342 \\
LR & 0.6861$\pm$0.0206 & 0.6882$\pm$0.0205 & 0.8534$\pm$0.0113 & 0.7416$\pm$0.0168 & 0.3212$\pm$0.0657 / 0.8876$\pm$0.0477 \\
RF & 0.6952$\pm$0.0319 & 0.7041$\pm$0.0284 & 0.8680$\pm$0.0146 & 0.7667$\pm$0.0228 & 0.3650$\pm$0.0330 / 0.8976$\pm$0.0102 \\
SVM & 0.6875$\pm$0.0229 & 0.6935$\pm$0.0193 & 0.8606$\pm$0.0081 & 0.7579$\pm$0.0097 & 0.2842$\pm$0.0264 / 0.9132$\pm$0.0177 \\
XGB & 0.7187$\pm$0.0258 & \textbf{0.7237$\pm$0.0240} & 0.8696$\pm$0.0153 & 0.7686$\pm$0.0280 & 0.3495$\pm$0.0861 / 0.8921$\pm$0.0321 \\
MLP-tabular & 0.6786$\pm$0.0300 & 0.6859$\pm$0.0282 & 0.8512$\pm$0.0171 & 0.7385$\pm$0.0221 & 0.2814$\pm$0.0408 / 0.9047$\pm$0.0188 \\
FT-Transformer & 0.6537$\pm$0.0150 & 0.6597$\pm$0.0119 & 0.8231$\pm$0.0162 & 0.6946$\pm$0.0283 & 0.2433$\pm$0.1014 / 0.8861$\pm$0.0568 \\
\midrule
Image-only & 0.4478$\pm$0.0239 & 0.4445$\pm$0.0212 & 0.6265$\pm$0.0118 & 0.4553$\pm$0.0189 & 0.0489$\pm$0.0257 / 0.9510$\pm$0.0411 \\
Scale-only & 0.6533$\pm$0.0195 & 0.6579$\pm$0.0144 & 0.8401$\pm$0.0129 & 0.7141$\pm$0.0178 & 0.3116$\pm$0.0768 / 0.8878$\pm$0.0392 \\
Late fusion (image + scale) & 0.6494$\pm$0.0195 & 0.6544$\pm$0.0157 & 0.8357$\pm$0.0156 & 0.7047$\pm$0.0220 & 0.2936$\pm$0.0567 / 0.8954$\pm$0.0290 \\
Concat fusion & 0.6679$\pm$0.0289 & 0.6743$\pm$0.0258 & 0.8346$\pm$0.0184 & 0.7093$\pm$0.0321 & 0.2650$\pm$0.0570 / 0.9001$\pm$0.0343 \\
Concat fusion + modality dropout & 0.6807$\pm$0.0391 & 0.6854$\pm$0.0356 & 0.8422$\pm$0.0210 & 0.7281$\pm$0.0265 & 0.2550$\pm$0.0491 / 0.9177$\pm$0.0222 \\
Missing-aware gated fusion & 0.6778$\pm$0.0261 & 0.6813$\pm$0.0270 & 0.8409$\pm$0.0131 & 0.7217$\pm$0.0242 & 0.2882$\pm$0.0673 / 0.8850$\pm$0.0181 \\
\midrule
Symmetric XAttn & 0.7139$\pm$0.0422 & 0.7187$\pm$0.0415 & \textbf{0.8737$\pm$0.0191} & 0.7807$\pm$0.0274 & \textbf{0.3728$\pm$0.0986 / 0.9033$\pm$0.0424} \\
AT-Attn (asymmetric + shortcut) & \textbf{0.7192$\pm$0.0242} & 0.7211$\pm$0.0225 & 0.8733$\pm$0.0130 & \textbf{0.7825$\pm$0.0178} & 0.3687$\pm$0.0862 / 0.8949$\pm$0.0293 \\
\bottomrule
\end{tabular}
\end{table*}

\textbf{Preprocessing and sequence construction.} T1-weighted MRI underwent a standard five-step preprocessing pipeline: DICOM-to-NIfTI conversion, affine registration to MNI152 space, skull stripping, N4 bias correction with z-score intensity normalization, and resampling to $96 \times 112 \times 96$ voxels at 2\,mm isotropic resolution. MRI records were aligned to the clinical visit grid by VISCODE normalization and exact subject-visit matching, and longitudinal trajectories were harmonized onto a shared visit grid spanning bl to m72, yielding a maximum sequence length of $T=9$; shorter sequences were zero-padded. Fig.~\ref{fig:data_panel}(B)--(C) illustrates the shared visit grid, representative multimodal missingness patterns, and the patient-level staging setup. For the cognitive-scale branch, missing scale values were forward-filled within each subject and any remaining entries were imputed by feature-wise medians computed from the training split. Missing MRI tensors were zero-filled only for batching and tensor-shape alignment, while downstream MRI-related modules used the MRI availability mask to exclude unmatched visits from contributing MRI-derived signals.

\textbf{Evaluation and training.} Evaluation uses patient-level stratified five-fold cross-validation, with each training fold further split into training and validation subsets to avoid patient leakage. For MCI Sens@90Spe, the operating threshold was selected on the validation fold to target 90\% specificity and then fixed on the corresponding test fold, where we report sensitivity together with the achieved specificity. The image backbone is a 3D ResNet-18 initialized with Med3D~\cite{chen2019med3d} pretrained weights. We use a unified feature dimension $D=256$ and multi-head attention with $H=4$ heads. Optimization uses AdamW with differential learning rates ($10^{-4}$ for the CNN backbone and $10^{-3}$ for the remaining modules), cosine annealing with 5-epoch linear warmup, and early stopping with patience 20 based on validation macro F1. Batch size is 8 on a single NVIDIA A100 80\,GB PCIe GPU.

\subsection{Comparison with Baselines}

Table~\ref{tab:main} summarizes five-fold cross-validation results. For fair comparison on the same patient-wise splits, all tabular baselines use the same patient-level representation, consisting of visit-level scale features, timestamps, visit-presence masks, and seven static variables. This representation is shared by classical machine-learning baselines (LR/SVM/RF/XGB/LGBM) and neural tabular baselines (MLP-tabular and FT-Transformer). Performance is evaluated by accuracy, macro F1, ROC-AUC, PR-AUC, and MCI Sens@90Spe.

The Image-only baseline is markedly weaker than baselines using cognitive-scale information, while the Scale-only baseline remains strong. This pattern suggests that, in the present cohort, MRI is not the primary decision source but a complementary modality whose value depends on how it is integrated.

Among the multimodal baselines, simple fusion strategies remain limited: plain concatenation is weak, and neither modality dropout nor missing-aware gated fusion closes the gap to AT-Attn. This pattern suggests that the gain of AT-Attn does not come from merely adding MRI, but from a more stable and structured way of injecting MRI information into a stronger clinical stream. Relative to the Scale-only and Concat baselines, AT-Attn improves all major evaluation metrics.

The comparison with tabular baselines is more balanced. Tree-based models are the strongest tabular references, with XGB achieving the best Macro-F1. For other metrics, the best results are obtained by AT-Attn variants: the asymmetric model gives the highest accuracy and PR-AUC, while the symmetric variant gives the highest ROC-AUC and MCI Sens@90Spe. Because the two variants are otherwise close, we use asymmetric AT-Attn as the main model for its better overall balance and its one-way guidance from cognitive scales to MRI features.

\subsection{Ablation Studies}

\begin{table*}[!tb]
\centering
\caption{ABLATION RESULTS OF AT-Attn ON THE ADNI COHORT UNDER FIVE-FOLD CROSS-VALIDATION. THE BEST RESULTS ARE IN BOLD.}
\label{tab:ablation}
\setlength{\tabcolsep}{4pt}
\small
\begin{tabular}{@{}lccccc@{}}
\toprule
Configuration & Acc & Macro-F1 & ROC-AUC & PR-AUC & MCI Sens@90Spe / Achieved Spe \\
\midrule
AT-Attn full (asym. + shortcut) & \textbf{0.7192$\pm$0.0242} & \textbf{0.7211$\pm$0.0225} & 0.8733$\pm$0.0130 & \textbf{0.7825$\pm$0.0178} & 0.3687$\pm$0.0862 / 0.8949$\pm$0.0293 \\
\midrule
\multicolumn{6}{@{}l}{\textit{Cross-attention direction}} \\
Symmetric XAttn & 0.7139$\pm$0.0422 & 0.7187$\pm$0.0415 & \textbf{0.8737$\pm$0.0191} & 0.7807$\pm$0.0274 & 0.3728$\pm$0.0986 / 0.9033$\pm$0.0424 \\
\midrule
\multicolumn{6}{@{}l}{\textit{Temporal modeling}} \\
No time bias & 0.6983$\pm$0.0281 & 0.7036$\pm$0.0280 & 0.8681$\pm$0.0122 & 0.7717$\pm$0.0153 & \textbf{0.3959$\pm$0.0293 / 0.8796$\pm$0.0300} \\
Fixed $\tau$ & 0.7099$\pm$0.0271 & 0.7124$\pm$0.0255 & 0.8721$\pm$0.0097 & 0.7757$\pm$0.0146 & 0.3668$\pm$0.0328 / 0.9115$\pm$0.0121 \\
Timestamp-shuffled & 0.6656$\pm$0.0368 & 0.6713$\pm$0.0383 & 0.8432$\pm$0.0238 & 0.7318$\pm$0.0381 & 0.2975$\pm$0.0706 / 0.8831$\pm$0.0174 \\
\midrule
\multicolumn{6}{@{}l}{\textit{Architecture and fusion components}} \\
w/o Change-and-Time & 0.6759$\pm$0.0187 & 0.6809$\pm$0.0122 & 0.8435$\pm$0.0168 & 0.7231$\pm$0.0267 & 0.3553$\pm$0.0803 / 0.8750$\pm$0.0348 \\
w/o shortcut & 0.7038$\pm$0.0335 & 0.7063$\pm$0.0334 & 0.8594$\pm$0.0195 & 0.7586$\pm$0.0297 & 0.2971$\pm$0.0692 / 0.9181$\pm$0.0253 \\
w/o gate (add) & 0.6878$\pm$0.0235 & 0.6908$\pm$0.0267 & 0.8666$\pm$0.0115 & 0.7659$\pm$0.0181 & 0.3244$\pm$0.1026 / 0.9061$\pm$0.0377 \\
\midrule
\multicolumn{6}{@{}l}{\textit{Input contribution controls}} \\
w/o static features & 0.6997$\pm$0.0148 & 0.7042$\pm$0.0146 & 0.8684$\pm$0.0138 & 0.7707$\pm$0.0216 & 0.3585$\pm$0.0658 / 0.8943$\pm$0.0243 \\
Clinical-only matched AT-Attn & 0.6935$\pm$0.0228 & 0.6982$\pm$0.0172 & 0.8596$\pm$0.0208 & 0.7582$\pm$0.0318 & 0.3491$\pm$0.0384 / 0.8967$\pm$0.0304 \\
\bottomrule
\end{tabular}
\end{table*}

Table~\ref{tab:ablation} reports the ablations grouped by design factor. All variants use the same patient-level split and training protocol as the full model unless otherwise stated.

\textbf{Impact of cross-attention direction.} Symmetric cross-attention remains close to the full asymmetric model: it slightly improves ROC-AUC and MCI Sens@90Spe, but is worse on accuracy, macro F1, and PR-AUC. We therefore interpret it as an operating-point trade-off, with asymmetric AT-Attn serving as the main configuration for balanced overall performance and one-way guidance from cognitive scales to MRI features.

\textbf{Impact of temporal modeling.} Although attention feature compatibility remains informative without an explicit time bias, removing the bias reduces accuracy and macro F1, and fixing $\tau$ gives smaller but consistent declines. Thus, the learned exponential bias offers the best overall balance; the higher MCI Sens@90Spe of the no-bias variant appears to be a threshold-specific trade-off. Timestamp shuffling, while preserving the underlying feature values, causes a broader degradation than either time-bias variant, suggesting that AT-Attn benefits from true chronological alignment rather than scale content alone. Because this drop exceeds that of the no-bias variant, the gain is unlikely to reflect a simple recency prior alone and instead depends on temporally consistent structure across Change-and-Time encoding, cross-attention, and pooling.

\textbf{Impact of architecture and fusion components.} Removing Change-and-Time information causes the strongest component-level degradation, including clear drops in macro F1 and PR-AUC, indicating the importance of time-aware change modeling. Removing shortcut paths lowers performance, confirming that direct unimodal evidence should remain visible to the classifier rather than being forced entirely through cross-attention. Replacing gated fusion with direct addition also weakens performance, supporting adaptive gating as a better mechanism for balancing raw MRI evidence with cross-modal refinement.

\textbf{Impact of input contribution controls.} Removing static features yields a smaller but consistent decline, indicating that demographic and genetic variables provide complementary context. The matched clinical-only AT-Attn variant preserves the same architecture and training protocol but masks the MRI branch; its lower aggregate performance relative to the full model suggests that MRI adds complementary information beyond longitudinal cognitive-scale and static clinical variables.

\subsection{Subgroup Gain Analysis}

\begin{figure}[!t]
\centering
\includegraphics[width=\columnwidth]{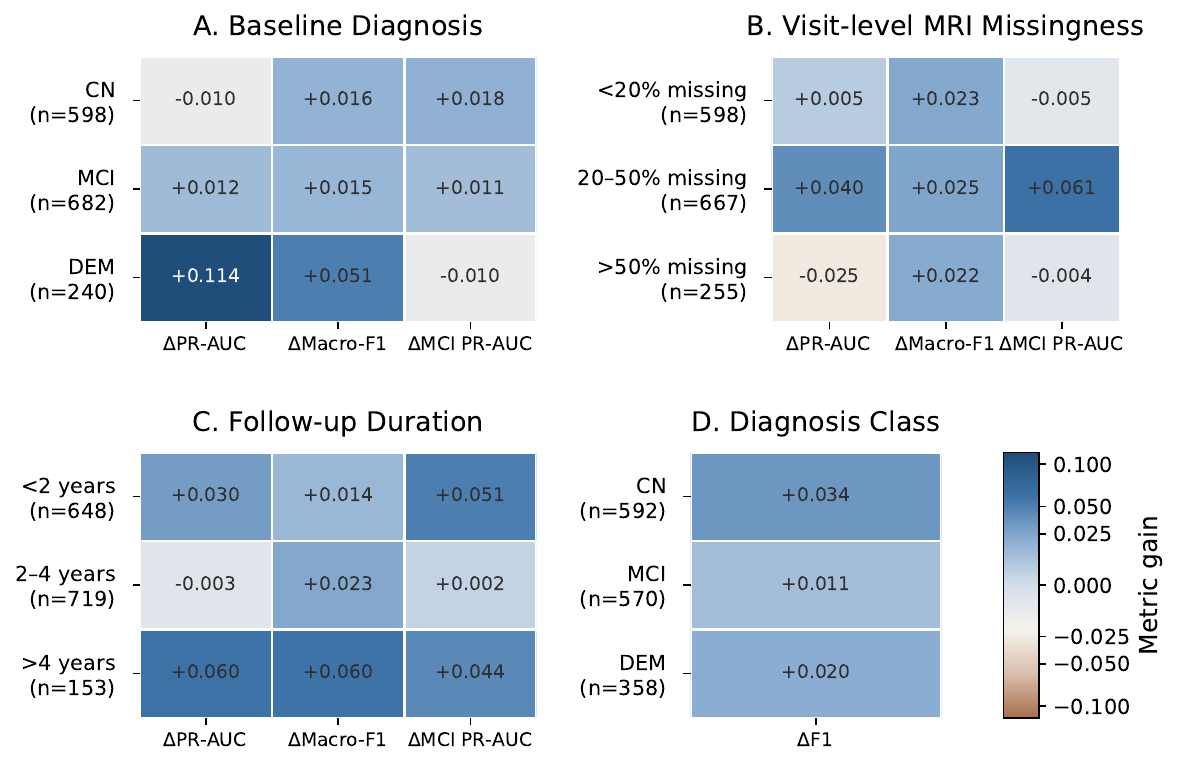}
\caption{Subgroup gain analysis of the full AT-Attn model relative to the matched clinical-only variant. Panel (A) summarizes gains in $\Delta$PR-AUC, $\Delta$Macro-F1, and $\Delta$MCI PR-AUC by baseline diagnosis. Panel (B) summarizes the same metrics by visit-level MRI missingness, and panel (C) by follow-up duration. Panel (D) shows class-wise $\Delta$F1. Positive values favor AT-Attn.}
\label{fig:subgroup_gain}
\end{figure}

Fig.~\ref{fig:subgroup_gain} suggests that the contribution of MRI is not uniform across patients, but context-dependent. In panel~(A), the clearest gains appear when the retained longitudinal sequence starts in the DEM group, especially in PR-AUC and Macro-F1, suggesting that MRI provides additional structural information when cognitive impairment is already substantial. This interpretation is clinically plausible and consistent with biomarker models suggesting that structural MRI, as a downstream marker of neurodegeneration, becomes more abnormal and clinically informative as Alzheimer's disease progresses~\cite{jack2013tracking}. Panels~(B) and~(C) further show that gains are more evident under moderate visit-level MRI missingness and longer follow-up, supporting MRI as a context-dependent complementary modality. Panel~(D) shows that class-wise $\Delta$F1 remains positive for all three diagnostic classes, with the largest improvement observed for CN.

\subsection{Attention Pattern Analysis}

\begin{figure*}[!tb]
\centering
\includegraphics[width=0.98\textwidth]{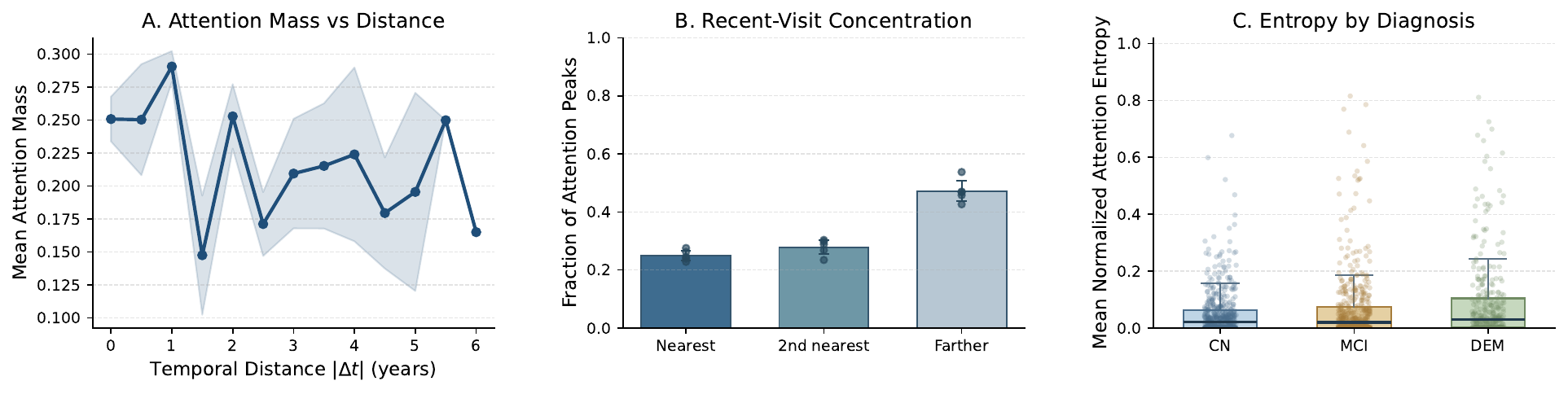}
\caption{Cohort-level attention analysis of the full AT-Attn model across the five held-out test folds. (A) Mean attention mass by temporal distance $|\Delta t|$ between MRI query and cognitive key/value visits (95\% CI across folds). (B) Fraction of attention rows whose peak falls on the nearest, second-nearest, or a more distant visit. (C) Subject-level mean normalized attention entropy by final diagnosis.}
\label{fig:attn_cohort}
\end{figure*}

Fig.~\ref{fig:attn_cohort} summarizes cohort-level attention behavior across the five held-out folds. Attention is not restricted to temporally nearest visits: although shorter gaps often receive higher mass, substantial attention is still assigned to more distant visits, and the nearest visit accounts for only about one quarter of attention peaks, whereas farther visits account for nearly half. This indicates that AT-Attn uses temporal distance as a soft preference rather than a hard nearest-visit rule. Subject-level attention entropy is slightly higher in MCI and DEM than in CN, but the overlap across groups suggests that this pattern is descriptive rather than strongly diagnosis-specific.

\begin{table}[!t]
\centering
\caption{Representative corrected cases. ``FU / MRI'' denotes follow-up duration and the number of MRI-available visits over all visits. The final column lists the top-2 MRI visits contributing to the final decision.}
\label{tab:case_corrections}
\small
\setlength{\tabcolsep}{3pt}
\begin{tabular}{@{}lccccp{0.31\columnwidth}@{}}
\toprule
\multirow[t]{2}{*}{Case} & \multirow[t]{2}{*}{True} & \multirow[t]{2}{*}{Clin-only} & \multirow[t]{2}{*}{AT-Attn} & \multirow[t]{2}{*}{FU / MRI} & Top-2 MRI visits \\
 &  &  &  &  & (year, weight) \\
\midrule
\multirow[t]{2}{*}{A} & \multirow[t]{2}{*}{CN} & \multirow[t]{2}{*}{MCI} & \multirow[t]{2}{*}{CN} & \multirow[t]{2}{*}{4.0 y; 3/6} & 2.0y (0.638) \\
 &  &  &  &  & 4.0y (0.326) \\
\multirow[t]{2}{*}{B} & \multirow[t]{2}{*}{MCI} & \multirow[t]{2}{*}{CN} & \multirow[t]{2}{*}{MCI} & \multirow[t]{2}{*}{4.0 y; 3/4} & 0.0y (0.501) \\
 &  &  &  &  & 3.0y (0.250) \\
\multirow[t]{2}{*}{C} & \multirow[t]{2}{*}{DEM} & \multirow[t]{2}{*}{MCI} & \multirow[t]{2}{*}{DEM} & \multirow[t]{2}{*}{2.0 y; 3/3} & 1.0y (0.514) \\
 &  &  &  &  & 0.0y (0.321) \\
\bottomrule
\end{tabular}
\end{table}

Table~\ref{tab:case_corrections} complements the cohort-level summary with three representative corrected cases. In all three cases, the clinical-only model misclassified the patient, whereas AT-Attn recovered the correct label after incorporating longitudinal MRI information. The top contributing MRI visits were not always the most recent ones, suggesting that earlier MRI can still provide useful evidence when the clinical trajectory is ambiguous.

\subsection{Discussion}

\textbf{Clinical interpretation.} The results suggest that structural MRI provides complementary diagnosis-support value when combined with longitudinal cognitive and static clinical information. In the present cohort, MRI is not the primary decision source; rather, its value depends on how it is integrated with stronger clinical signals. The multimodal comparisons indicate that simply adding MRI is insufficient, whereas AT-Attn provides a controlled fusion strategy that allows MRI to contribute without overwhelming direct clinical evidence. The subgroup and case analyses further suggest that this contribution is context-dependent rather than uniform across patients. Taken together, these findings support the use of structural MRI as complementary evidence for patient-level diagnosis support when it is incorporated through a stable and temporally informed fusion mechanism.

\textbf{Limitations.} Several limitations remain. First, because ADNI clinical diagnoses are strongly tied to cognitive and functional assessments, the longitudinal cognitive-scale inputs may partly overlap with information used in label assignment. In addition, the current task is best interpreted as multimodal diagnosis support for patient-level staging rather than fully independent biomarker-based prediction or future conversion forecasting. Future work should evaluate stricter settings, such as excluding diagnosis-defining scales and predicting future conversion from earlier visits only. Second, although five-fold cross-validation improves robustness over a single split, the study is still limited to ADNI and requires external-cohort validation. Finally, the current formulation assumes a shared visit timeline across modalities and therefore does not yet handle fully asynchronous MRI and cognitive sampling.

\section{Conclusion}
\label{sec:conclusion}

We present AT-Attn, a temporal-aware multimodal framework for longitudinal AD diagnosis support. By combining Change-and-Time encoding, time-biased asymmetric cross-attention, and gated fusion, the framework integrates MRI with longitudinal clinical information in a stable and constrained manner. On ADNI, AT-Attn outperforms unimodal and naive fusion baselines while remaining competitive with strong tabular baselines. Overall, the results suggest that structural MRI provides context-dependent complementary value when incorporated through controlled fusion rather than naive multimodal combination. Future work will extend this evaluation to stricter diagnosis-support settings, asynchronous multimodal timelines, and external validation cohorts.

\balance
\bibliographystyle{IEEEtran}

\end{document}